# Computing Fuzzy Rough Set based Similarities with Fuzzy Inference and Its Application to Sentence Similarity Computations

Nidhika Yadav[a]
Ph.D., IIT Delhi

**Abstract-** *Several research initiatives have been proposed for computing similarity between two Fuzzy Sets in analysis through Fuzzy Rough Sets. These techniques yield two measures viz. lower similarity and upper similarity. While in most applications only one entity is useful to further analysis and for drawing conclusions. The aim of this paper is to propose novel technique to combine Fuzzy Rough Set based lower similarity and upper similarity using Fuzzy Inference Engine. Further, the proposed approach is applied to the problem computing sentence similarity and have been evaluated on SICK2014 dataset.*

## 1. Introduction

The foundations of theory of Rough Set were laid by Pawlak (Pawlak, 1982). Since then, it has been of interest both for theoretical researcher and application scientists for applications varying from financial analysis, text summarization, image processing, information retrieval, stock prediction to keyword extraction, feature selection to mention a few. Fuzzy Rough Set (Pawlak, 1982, Jensen and Shen, 2004) was proposed for fuzzification of lower and upper approximation, which in applications to problems is very intuitive and useful, given the fact that each member of the universe bears a membership to the set under consideration. Since, there are two sets in a Rough Set based application namely lower approximation and upper approximation, hence, there are two Fuzzy Sets under study in the domains of Fuzzy Rough Set based analysis.

In problems in which two Fuzzy Rough Sets have been computed and compared, how to determine the similarity and other relations between these pairs of Fuzzy Rough Sets? This can be illustrated with the example of application validated in this paper, viz. computing sentence similarities. Consider a universe of words in the context under consideration. For each sentence can be represented as a Fuzzy Rough Set, in terms of membership of each word in universe to the sentence. Two Fuzzy Sets correspond to the lower approximation and upper approximation respectively. Hence, each of the two sentences are represented as a pair these two Fuzzy Sets. To compare similarity between two sentences. We have another pair of Fuzzy Sets arising from second sentence. Two Fuzzy Rough Sets based similarities can be computed for a generic use in a variety of tasks not limiting to sentence semantic similarities. These include any Fuzzy

[a] Email Address: nidhikayadav678@outlook.com

Rough Set based application which include comparison of two objects. This can be image recognition problem to edge detection application.

Text semantic similarity is the task in which two text fragments are given and a similarity score among them is required to be evaluated. The fragments can be documents on whole or just two phrases of text and may be two sentences. The applications of semantic similarity computations are varied, one major application that it can trigger in near future is comment analytics. The ability of an organization or individual to analyse his or her comments for the statistics not just in two class classification of number of likes or dislikes but also in terms of similarity of post with the comments and comments among each other to draw conclusion of a imperative output of the post. This post can be on Twitter, Facebook, LinkedIn or any social media platform. This indeed would help the businesses, groups and individuals with huge followers to get the statistics both quantitative as well as semantics for a quick analysis of the post. Other applications involve deep semantic application which include question answering, text summarization, keyword extraction to mention a few.

In this paper we shall be considering text sentence similarity using the proposed technique of combining the two similarities viz. lower similarity and upper similarity into one quantity, using Fuzzy Inference Engine. This is a logical and mathematically well-established proposal. Further, we shall also be analysing the unification of these two similarity measures into one unit evaluation in terms of mean square error for performance on SICK2014 dataset.

The paper is organised as follows. Section 2 briefly discuss previous works in these topics. The proposed technique is presented in Section 3. Section 4 gives results and Section 5 proposes future works in this area.

## 2. Previous Work

Rough Set and Fuzzy Rough Set based similarities dates back to inception of Rough Sets. Das-Gupta (1988) defined Rough Set based lower similarity and Rough Set based upper similarities. Here the authors focussed on query terms and document similarity. Further, at that time no considerable computational models for evaluating semantic similarities were prevalent in applications. Various applications of Fuzzy Rough Set based sentence similarity measure have been studied till now. Fuzzy Rough Sets deal with fuzzification of Rough Sets. For this we here give a brief overview of Fuzzy Sets, Rough Sets and Fuzzy Rough Sets.

Given a universe U, in a crisp set an object belongs to the set or it does not. While in Fuzzy Sets, each element of universe belongs to a set with a particular membership. The membership values vary form zero through 0.5 to one. Here, a membership of 0.9 means the element of universe belongs to the set under consideration with a degree of 0.9, and may be some other set too with other membership value. Rough Sets on the other hand deals with partial or complete belonginess of concepts present in knowledge granules. Lower approximation of a set X consists of all those elements of universe which are contained in the set X given a knowledge R about objects. The upper approximation of a set X consists of all those elements of universe whose knowledge granules have some points in common with X.

A fuzzy equivalence relation on universe U is a relation that satisfies reflexivity, symmetry and transitivity. Given a Fuzzy equivalence relation R, and a membership computation method μ, which computes how much an element belongs to the set under consideration, X . Fuzzy-Rough lower approximation and the Fuzzy-Rough upper approximations of set X are given as follows (Das-Gupta, 1988, Srinivasan, 2001):

$$\underline{R}(X)(x) = \min\{\max(1 - R(x,y), \mu(y)): y \in U\}$$

$$\overline{R}(X)(x) = \max\{\min(R(x,y), \mu(y)): y \in U\}$$

In, Chatterjee and Yadav (2015), $R(x,y)$ is computed using gaussian function where the gaussian parameters are learnt via PSO algorithm. While in this paper we shall be computing it using WuPalmer similarity measure (Wu and Palmer, 1994).

Singh and Dey (2005) used Fuzzy Rough Set for client-side user specific information retrieval. It is mentioned here as some relation exists between the problem, we are tackling in this paper viz. sentence similarity and the way Singh and Dey used query and document grading for information retrieval. However, their model was very restrictive to the constrained application, retrieval of documents on client side, while we in this paper present a general framework to be used in any application using Fuzzy Rough Sets.

Recently, Chatterjee and Yadav (2019) proposed the use of Fuzzy Rough Sets for text summarization wherein they compute lower and upper similarities. Some earlier works are from literature are in information retrieval which include Srinivasan (2001). Not much evaluations can be found in sentence similarity using Fuzzy Rough Sets in past works. However, in broader context Fuzzy Rough Sets have been used in several applications which include image processing (Namburu et al, 2017), classification (Hu et al, 2010), feature selection (Jensen and Shen, 2004), text summarization (Chatterjee and Yadav, 2019) and keyword extraction (Jensen and Shen, 2004), to mention key areas.

The next task is computing the similarities between two pair of Fuzzy Rough Sets. Consider another sets X and Y and the corresponding Fuzzy Rough Sets. The way to compute similarity between X and Y is to use the Fuzzy Rough Sets to compute the similarity between these sets, viz. $\underline{R}(X), \underline{R}(Y)$ and the pairs $\overline{R}(X), \overline{R}(Y)$. These inscribe in it the use of knowledge R and the membership computations, which can be computed using Rough Set based technique as in (Chatterjee and Yadav, 2019). The similarity measure between sets X and Y, given knowledge R, is given as follows (Das-Gupta, 1988):

$$\underline{\text{Sim}}(X, Y) = \frac{|\underline{R}(X) \cap \underline{R}(Y)|}{|\underline{R}(X) \cup \underline{R}(Y)|}$$

$$\overline{\text{Sim}}(X, Y) = \frac{|\overline{R}(X) \cap \overline{R}(Y)|}{|\overline{R}(X) \cup \overline{R}(Y)|}$$

Several other techniques can be used to compute similarities, some use cosine similarity, Euclidean distance, co-relation or even other distance metrics. In the following section, we shall discuss the proposed technique to compute sentence similarity between two text fragments or sentences in detail.

## 3. Proposed Technique

Sentence similarity on the other hand is pivotal for points discussed in introduction. We shall be working on SICK2014 dataset (Marelli et al, 2014) for computing the evaluations. This data set has two text fragments, the relatedness score between the textual data and entailment category assigned to the relation of first to the second. The proposed technique computes similarity between two sentences is based on Fuzzy Rough Sets and is described in details below.

Consider two sentences, S1 and S2, whose similarity need to be computed. Consider a universe, U, under consideration. This universe can be a superset of all words in union of S1 and S2 or an exact union. In case of superset the problem can be solved in much greater details as in more contextual information is present. Especially useful in disambiguating words and in problems such as machine translations. In the present experimentational work to propose and evaluate the efficacy of novel model to compute similarity of sentences, we shall use universe as exact union of words of importance, by removing trivial words. Once U, S1 and S2 have been selected. Fuzzy lower and upper approximations for sentence S1 are computed as follows.

$$\underline{R}(S1)(word1) = \min\{\max(1 - R(word1, word2), \mu(word2)): word2 \in U\}$$

$$\overline{R}(S1)(word1) = \max\{\min(R(word1, word2), \mu(word2)): word2 \in U\}$$

Chatterjee and Yadav (2019) have worked on using deep learning based embeddings in their work to use Fuzzy Rough Sets for computing sentence similarities. Further, they have used gaussian function to evaluate the relation or similarity between words x and y. The membership function there is computed using Rough Set based membership where information table is obtained using discretization of embedding vectors.

However, in this work, we shall not be using deep learning based embeddings to represent words. We have used wordnet, wordnet based similarities to compute R(x, y), extent of relation between words x and word y. Also, the membership function is computed using wordnet. The main reason is to keep model simple and focus on the main area of emphasis in the paper, instead of dwelling in details to improve performance further. The main aim of paper is to find mathematical sound and efficient way to get a right single numeric quantity as an output of similarity of two Fuzzy Rough Sets.

Consider sentence S1 and sentence S2, a knowledge R, which we have taken as wordnet and universe U under consideration. Then these two sentences S1 and S2, are considered as set theoretic sets of objects which are words in this application. Using Fuzzy Rough Sets, we get four sets, lower approximations of S1 and S2 and corresponding upper approximations. The four sets symbolically are $\underline{R}(S1), \underline{R}(S2)$, the lower approximations and the upper approximations, $\overline{R}(S1), \overline{R}(S2)$ respectively. One must note that these are all Fuzzy Sets. To compute similarity between S1 and S2 using these sets, with added information about the problem at hand, one can use any distance measure. We have illustrated the application using cosine similarity between corresponding sets, viz. similarity between Fuzzy Sets $\underline{R}(S1), \underline{R}(S2)$ and Fuzzy Sets $\overline{R}(S1), \overline{R}(S2)$. This gives us two similarity measures, lower similarity and upper similarity. Here, the main problem statement of the paper comes to focus. The aim to combine these two similarities to get one measure, called rank, which is logically correct and experimentally validated. For this we have designed a Mamdani Fuzzy Inference System. Once the final rank is computed between the two sets, we compare the relatedness of this rank with the relatedness score given to us in SICK2014 dataset for validations. The Mamdani Fuzzy Inference System is described below.

### 3.1. Rank Score using Mamdani Fuzzy Inference System

The evaluation of sentence similarity between two sentences using Fuzzy Rough Sets is computed as described above. Hence, for each pair of lower and upper approximation for each of the two sentences, we have a pair of similarities, the lower similarity and the upper similarity. These similarities do not give a concise measure of final output to be project to a user or another application to consume it. Hence, a Mamdani Fuzzy Inference System is designed to get the final rank based on these two-similarity index we generate from an input of two sentences.

The Mamdani Fuzzy Inference System (FIS) is described as follows:

1. Inputs: The two inputs to the FIS are:
   a. Lower Similarity Score: This is obtained by computing the similarity scores from lower approximation of sentence 1 and lower approximation of sentence 2.
   b. Upper Similarity Score: The upper similarity score is obtained from similarity computation from upper approximation of sentence 2 and upper approximation of sentence 2.

   Both these similarity scores are measured as three linguistic Fuzzy Sets viz, low, average and high.

2. Output Rank. The output is the similarity rank in range 0 to 5. This is measured as low, medium and high. Each a Fuzzy Set.

The Fuzzy Sets were learned from grid-based selection and comparison of mean square error between the final output as presented by language experts. he best Fuzzy Modeling of the two linguistic inputs and output are presented in Fig 1, Fig 2 and Fig 3 respectively. Fig 4 shows computation of final relatedness rank using defuzzification process of FIS for a sample input.

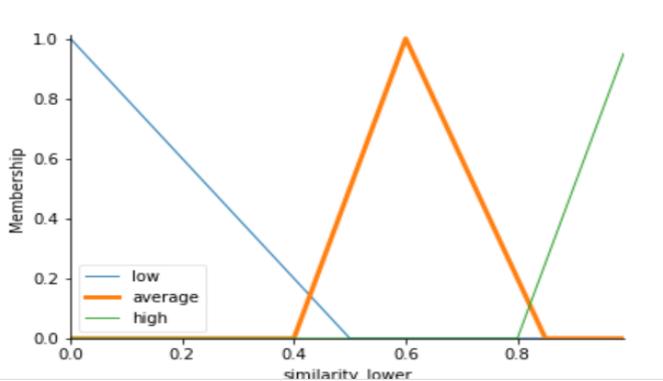
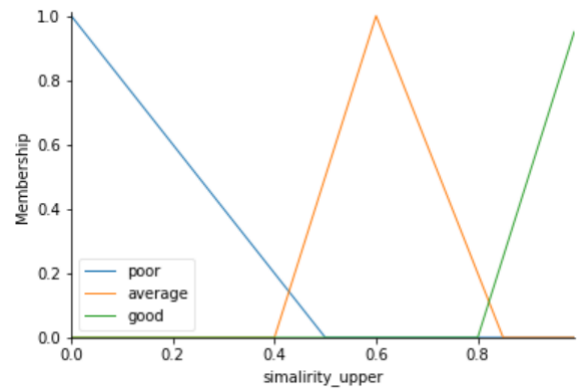

Fig 1. The Fuzzy Sets for Lower Similarity    Fig 2. The Fuzzy Sets for Upper Similarity

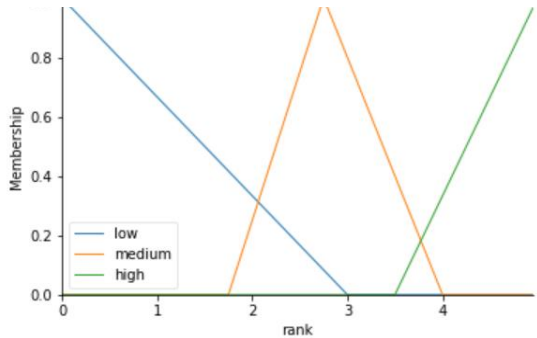 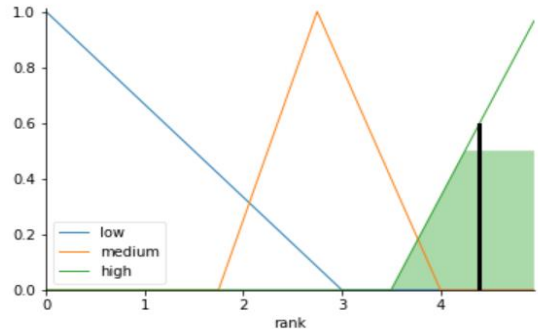

Fig 3. The Fuzzy Sets for Similarity Rank in range 0 to 5

Fig 4. Computation of final rank using FIS

The proposed computation model has been evaluated on SICK2014 dataset. SICK2014 stands for Sentence Involving Compositional Knowledge, developed in 2014 for sematic relatedness computations. SemEval-2014 task considers evaluation of Compositional Distributional Semantic Models on Full Sentences through Semantic Relatedness and Textual Entailment.

The SICK2014 dataset has been constructed for distributional semantic models, which analyse the distributional view of terms and concepts in a text. How terms are distributed in a text fragment lays emphasis on its meanings and context. Instances of knowledge concepts has been considered in the datasets. Knowledge such as the fact that woman and girl both represent female gender, for an example.

Two tasks have been considered here:

(i) Semantic Relatedness. Here a score is assigned for the relation between the two text fragments. How much are they talking of the same topic. Whether it is contradictory or not does not matter in this measure, just the context is measured for 10,000 sentences.
(ii) Textual Entailment. Here it the contradiction, implication and negation between the two text contexts has been collected for same 10,000 sentences.

The dataset has been developed by motivation of these two primary evaluation concerns. Prior to this semantic similarity and textual entailment datasets were present which utilized the use of external databases such as Encyclopaedia, for identifying named entities and n-gram words.

The key aim of this dataset development was analysing compositional properties of text, apart from providing a good mixture of sentences which can bring out sentence similarity and entailment projected via areas model techniques can grasp. The dataset consists of a header line and below it is the data, each line of data is tab separated and containing the following:

1. Sentence 1. First text fragment.
2. Sentence 2. Second text fragment.
3. Similarity score between Sentence 1 and Sentence 2. This score is measure in count from 0 to 5, 5 being the highest relatedness score.
4. The entailment category. Is one sentence implied by the other, whether the second sentence is true give the first, knowledge, or independent of each other or that the second sentence is false given the truth of first.

The datasets were added with entries containing negative of the submitted entries, in total making 10,000 with a right mix of all three categories of entailments. The dataset is dived into 50-50 ratio for training and testing. 21 teams participated in evaluations and around 19 submitted results. Results were measure in terms of mean square errors and co-relation coefficients. Most teams used external databases such as WordNet.

| Fuzzy Sets | Linguistic Variables | Type of Membership | The Parameters of Memberships Model 1 | The Parameters of Memberships Model 2 |
|---|---|---|---|---|
| similarity_lower | similarity_lower['low'] | trigular membership function | [0, 0 , 0.5] | [0, 0 , 0.5] |
|  | similarity_lower['average'] | trigular membership function | [0.4, 0.6, 0.85] | [0.4, 0.7, 0.95] |
|  | similarity_lower['high'] | trigular membership function | [0.8, 1, 1] | [0.9, 1, 1] |
|  |  |  |  |  |
| simalirity_upper | simalirity_upper['poor'] | trigular membership function | [0, 0 , 0.5] | [0, 0 , 0.5] |
|  | simalirity_upper['average'] | trigular membership function | [0.4, 0.6, 0.85] | [0.4, 0.7, 0.95] |
|  | simalirity_upper['good'] | trigular membership function | [0.8, 1, 1] | [0.9, 1, 1] |
|  |  |  |  |  |
| rank | rank['low'] | trigular membership function | [0, 0, 3] | [0, 0, 3] |
|  | rank['medium'] | trigular membership function | [1.75, 2.75, 4] | [1.75, 2.75, 4] |
|  | rank['high'] | trigular membership function | [3.5,5, 5] | [3.5,5, 5] |

Table 1. The details of the final Fuzzy System build and one of the five being tested.

```
rule1 = (similarity_lower['low'] & simalirity_upper['poor'], rank['low'])
rule2 = simalirity_upper['average'] & simalirity_upper['poor'], rank['medium'])
rule3 = (similarity_lower['high'] & simalirity_upper['average'], rank['high'])
rule4 = (similarity_lower['average'] & simalirity_upper['good'], rank['high'])
rule5 = (similarity_lower['average'] & simalirity_upper['average'], rank['medium'])
rule6 = (similarity_lower['high'] & simalirity_upper['good'], rank['high'])
rule7 = (similarity_lower['low'] & simalirity_upper['average'], rank['low'])
rule8 = (similarity_lower['low'] & simalirity_upper['good'], rank['medium'])
rule9 = (similarity_lower['high'] & simalirity_upper['poor'], rank['medium'])
```

Fig 5. Sample rules to compute the FIS.

In our experiments, training was performed on 500 randomly selected entries of data and grid-based values of Fuzzy Inference Engine were evaluated for testing the final FIS to be used on testing set. Further, Fuzzy Rough Sets relation used R(x, y) via Wu-Palmer (Wu and Palmer, 1994) similarity of Wordnet and membership function was taken as 1 in case of exact match and if not the maximum similarity of word with sentence, otherwise.

Table 1 shows the details of Fuzzy Inference Engine being experimented on, it includes the best one and one of the five other models which were evaluated. Figure 5 mentions the rules of FIS considered. Further, these rules changed with change in experiments. Table 2 gives sample intermediate computations of 7 out of 5000 testing combinations. It gives the rank assigned by expert with computed ranks by two of the five models, which include the best model.

The MSE obtained on the testing set of 500 sentences was **1.298** for the FIS depicted in Fig 1, 2 and 3. Further, this is near the participating team results of competition of SICK2014 Mean Square Error as can be seen in Table 7 of paper Marelli et al (2014).

| S.No. | Sentence 1 | Sentence 2 | Expert Rank | Lower Similarity | Upper Similarity | FIS Rank from First FIS | FIS Rank from Second FIS |
|---|---|---|---|---|---|---|---|
| 1 | A woman is wearing sunglasses of large size and is holding newspapers in both hands | There is no woman wearing sunglasses of large size and holding newspapers in both hands | 3.8 | 0.9308 | 0.9562 | 4.4285 | 4.3334 |
| 2 | A black dog and a small white and black dog are looking up at a kitchen countertop | A large dog and a small dog are standing next to the kitchen counter and are sniffing | 4.2 | 0.8129 | 0.8452 | 4.0414 | 2.8511 |
| 3 | A man is playing a guitar on stage | There is no man playing a guitar on stage | 3.6 | 0.9114 | 0.9407 | 3.8880 | 4.4056 |
| 4 | A man is playing a guitar on stage | A guitarist has blonde and flyaway hair | 3.0 | 0.7462 | 0.7698 | 2.8384 | 2.8562 |
| 5 | A man is playing a guitar on stage | A bald person is playing a guitar | 3.8 | 0.8832 | 0.9035 | 2.9939 | 4.3667 |
| 6 | An old woman is wearing a rose patterned shirt and is clumsily carrying two newspapers | There is no woman wearing sunglasses of large size and holding newspapers in both hands | 3.0 | 0.7522 | 0.7550 | 2.8531 | 2.8366 |
| 7 | Two girls are lying on the ground | Several children are sitting down and have their knees raised | 2.8 | 0.6962 | 0.7114 | 2.8335 | 2.8448 |

Table 2. The similarity ranks and expert ranks computed by two Fuzzy Inference Engines for sample 7 sentences in datasets was evaluated five times on random selection.

Hence, we conclude that firstly, we have combined two similarity scores into one using Mamdani Fuzzy Inference Engine over best learned Fuzzy Sets, chosen from grid search. An advantage of using FIS Ranker was the similarity be it cosine or any, ranges in 0 to 1, and FIS helped to model it to range 0, 5 and giving us one comprehensive value to analyse and evaluated. Further, a low score of MSE is a motivating factor. The next section discusses the future work and improvements possible in this area.

| Model | MSQ (Marelli et al, 2014) | Spearman Corelation |
|---|---|---|
| Proposed Model in this paper | 1.2980 | 0.3900 |
| asjai run5 | 1.1040 | 0.4610 |
| Yamraj_run1 | 2.6650 | 0.5360 |
| ASAP_run1 | 0.6620 | 0.5790 |
| StanfordNLP | 0.3230 | 0.7560 |

Table 3. Results compared with base paper best results.

Table 3 gives the result obtained on semantic similarity tasks compared with some of the best systems which participated in the sentence similarity task of SICK2014. Our results are not better than results reported in base paper Marelli et al, (2014). But, as we mentioned before this paper and work is an illustration of how the two similarities obtained by comparing two Fuzzy Rough Sets can be combined. And further optimizations are possible, but here our focus was to lay emphasis on use of Mamdani Fuzzy Inference in combining lower similarity and upper similarity into one unit, which can be used for further analysis and applications. Also, from this paper we lay emphasis on providing details of Fuzzy Engine in research papers, as this effects the results. These parameters of FIS learned by machine learning tools are needed for further optimised. Given the model was not optimised through parameter optimization, hence we conclude that the results are quite encouraging. Further, model as suggested by Chatterjee and Yadav (2019) can be combined in this framework to improve results more. Also, we computed one run of the system only. Hence, this is somewhat motivating proof of concept.

## 5. Future Work

The paper presents a way to combine the two similarity values into one using FIS Ranker, which produced MSE somewhat near to the reported results of base paper (Marelli et al, 2014). Its encouraging since it was a proof of concept model. The Fuzzy Rules needs to be automatically learned through the training data. Further, we suggest that changing a universe, which means the context, can have a major impact on the output produced. This fact can be analysed further, both theoretically and experimentally. This paper was for illustration of methodology and initial testing and validation of proof of concept, fine tuning and learning needs to be still done. Further, learning of Fuzzy Inference System and rules can be autogenerated using training as against hard coded grid-search to optimise the models further. Also, to improve results further word embeddings can be used and the similarity between words can be computed using deep learning models.

## References

1. Das-Gupta, P. (1988, May). Rough sets and information retrieval. In Proceedings of the 11th annual international ACM SIGIR conference on Research and development in information retrieval (pp. 567- 581), ACM.

2. Hu, Q., Zhang, L., Chen, D., Pedrycz, W., & Yu, D. (2010). Gaussian kernel based fuzzy rough sets: model, uncertainty measures and applications. International Journal of Approximate Reasoning, 51(4), 453-471.
3. Jensen, R., & Shen, Q. (2004). Fuzzy–rough attribute reduction with application to web categorization. Fuzzy Sets and systems, 141(3), pp. 469-485.
4. Jensen, R., & Shen, Q. (2007). Fuzzy-rough sets assisted attribute selection. IEEE Transactions on fuzzy systems, 15(1), pp. 73-89.
5. Marelli, M., Bentivogli, L., Baroni, M., Bernardi, R., Menini, S., & Zamparelli, R. (2014). Semeval-2014 task 1: Evaluation of compositional distributional semantic models on full sentences through semantic relatedness and textual entailment. In Proceedings of the 8th international workshop on semantic evaluation (SemEval 2014) (pp. 1-8).
6. Namburu, A., kumar Samay, S., & Edara, S. R. (2017). Soft fuzzy rough set-based MR brain image segmentation. Applied Soft Computing, 54, 456-466.
7. Pawlak, Z. (1982). Rough sets. International journal of computer & information sciences, 11(5), pp. 341- 356.
8. Pawlak, Z. (2012). Rough sets: Theoretical aspects of reasoning about data (Vol. 9). Springer Science & Business Medi.
9. Singh, S., & Dey, L. (2005). A rough-fuzzy document grading system for customized text information retrieval. Information processing & management, 41(2), pp. 195-216.
10. Srinivasan, P., Ruiz, M. E., Kraft, D. H., & Chen, J. (2001). Vocabulary mining for information retrieval: rough sets and Fuzzy Sets. Information Processing & Management, 37(1), pp. 15-38.
11. Wu, Z., & Palmer, M. (1994, June). Verbs semantics and lexical selection. In Proceedings of the 32nd annual meeting on Association for Computational Linguistics (pp. 133-138). Association for Computational Linguistics.